\documentclass[conference]{IEEEtran}
\IEEEoverridecommandlockouts
\usepackage{cite}
\usepackage{amsmath,amssymb,amsfonts}
\usepackage{algorithmic}
\usepackage{graphicx}
\usepackage{textcomp}
\usepackage{mathtools}
\usepackage{commath}
\usepackage{xcolor}
\graphicspath{ {./images/} }
\def\BibTeX{{\rm B\kern-.05em{\sc i\kern-.025em b}\kern-.08em
    T\kern-.1667em\lower.7ex\hbox{E}\kern-.125emX}}
\begin{document}

\title{A Daily Tourism Demand Prediction Framework Based on Multi-head Attention CNN: The Case of The Foreign Entrant in South Korea\\
{\footnotesize \textsuperscript{}}
\thanks{}
}

\makeatletter
\newcommand{\linebreakand}{%
  \end{@IEEEauthorhalign}
  \hfill\mbox{}\par
  \mbox{}\hfill\begin{@IEEEauthorhalign}
}
\makeatother

%\author{\IEEEauthorblockN{Anonymous Author}
%\IEEEauthorblockA{\textit{----} \\
%\textit{----}\\
%----, ---- \\
%----}}

\author{\IEEEauthorblockN{Dong-Keon Kim}
\IEEEauthorblockA{\textit{College of Computing} \\
\textit{Sungkyunkwan University}\\
Suwon, South Korea \\
kdk1996@skku.edu}
\and
\IEEEauthorblockN{Sung Kuk Shyn}
\IEEEauthorblockA{\textit{Department of Artificial Intelligence} \\
\textit{Sungkyunkwan University}\\
Suwon, South Korea \\
davidshyn@skku.edu}
\and
\IEEEauthorblockN{Donghee Kim}
\IEEEauthorblockA{\textit{College of Computing} \\
\textit{Sungkyunkwan University}\\
Suwon, South Korea \\
ym.dhkim@skku.edu}
\linebreakand
\IEEEauthorblockN{Seungwoo Jang}
\IEEEauthorblockA{\textit{Department of Electrical and Computer Engineering} \\
\textit{Sungkyunkwan University}\\
Suwon, South Korea \\
sewo@skku.edu}
\and
\IEEEauthorblockN{Kwangsu Kim}
\IEEEauthorblockA{\textit{College of Computing} \\
\textit{Sungkyunkwan University}\\
Suwon, South Korea \\
kim.kwangsu@skku.edu}
}

\maketitle

\begin{abstract}

Developing an accurate tourism forecasting model is essential for making desirable policy decisions for tourism management. Early studies on tourism management focus on discovering external factors related to tourism demand. Recent studies utilize deep learning in demand forecasting along with these external factors. They mainly use recursive neural network models such as LSTM and RNN for their frameworks. However, these models are not suitable for use in forecasting tourism demand. This is because tourism demand is strongly affected by changes in various external factors, and recursive neural network models have limitations in handling these multivariate inputs. We propose a multi-head attention CNN model (MHAC) for addressing these limitations. The MHAC uses 1D-convolutional neural network to analyze temporal patterns and the attention mechanism to reflect correlations between input variables. This model makes it possible to extract spatiotemporal characteristics from time-series data of various variables. We apply our forecasting framework to predict inbound tourist changes in South Korea by considering external factors such as politics, disease, season, and attraction of Korean culture. The performance results of extensive experiments show that our method outperforms other deep-learning-based prediction frameworks in South Korea tourism forecasting.

\end{abstract}


\section{Introduction}
As exchanges between countries increase, the tourism industry in each country becomes more important. For developing the tourism industry, tourism demand is essential in establishing tourism policy, business plan, and strategy revision. Therefore, tourism management needs to discover external factors related to tourism demand and design an accurate prediction model.

%Early studies related to tourism demand forecasting mainly propose a forecasting framework using statistical models such as ARIMA (AutoRegressive Integrated Moving Average) or SARIMA (Seasonal ARIMA) \cite{song2019review}. In particular, these studies mainly focus on discovering extrinsic factors related to specific tourism topics (e.g., accommodation demand, number of visitors to scenic spots \cite{dharmaratne1995forecasting}) in the specific region rather than developing forecasting models that improve accuracy for the prediction purpose itself. However, these models have a limitation in that their performance deteriorates in forecasting non-stationary time-series data such as tourism demand which has a certain period or trend.

Unlike earlier studies that mainly used regression models, recent tourism demand forecasting studies \cite{zhang2019forecasting, law2019tourism} utilize sequential deep learning models such as RNN (Recurrent Neural Network) \cite{rumelhart1986learning} and LSTM (Long Short Term Memory) \cite{hochreiter1997long} in their prediction framework. And the latest studies \cite{kulshrestha2020bayesian, lu2020method} of tourism demand forecasting mainly design a framework based on this sequential neural network model. These predictive models show better accuracy than regression models.

Recurrent-based networks (RNN, LSTM) are mainly used in tourism demand forecasting due to their excellent performance, but have the following limitations. First, the recurrent network models are structured to extract temporal features of a single variable, while tourism demand variables are affected by other external factors (e.g., the number of tourists entering a country is influenced by oil prices.) \cite{law2019tourism}. So, the recurrent models are difficult to interpret variable-wise correlation, which leads to the limitation of forecasting performance. In addition, RNN and LSTM models have an autoregressive structure in which prediction values are put back as input. This structure has poor long-term prediction accuracy when the data has a nonlinear trend, whereas the long-term forecasting of tourism demand plays a vital role from a practical point of view. Therefore, autoregressive-based prediction models are not suitable for practical forecasting.

We present a Multi-Head Attention Convolutional neural network (MHAC) model for forecasting tourism demand to address the problems mentioned above. The proposed model receives historical multivariate time-series information and outputs a sequence of how the interest variable will change in the future. With multivariate time series as inputs, separated CNN (Convolutional Neural Network) layers independently interpret temporal patterns of the individual variable. Also, the model employs an attention module for discovering correlations between multiple variables. Finally, the tourism demand prediction sequence is output at once through attention content and temporal feature.

The proposed method is designed to predict the demand for foreign entrants in South Korea. To the best of our knowledge, there are no deep-learning-based prediction framework studies suitable for South Korea's tourism data. We design a deep learning model that predicts tourism demand in South Korea using a multivariate time series. As in other tourism management studies \cite{song2019review}, we consider several extrinsic factors related to the inbound tourist of South Korea. Then, the proposed MHAC model predicts the tourist trend by reflecting the considered variables and the historical tourist trend data as inputs. As a result of the prediction experiment, our forecasting framework shows the high accuracy of demand forecasting for tourists visiting South Korea. Especially, it shows good forecasting performance even in extreme situations (e.g., COVID-19 pandemic).

Our contributions in this paper are as follows:
\begin{itemize}
  \item A novel time-series forecasting framework for tourism management is proposed. This framework has a structure that receives several extrinsic variables as input and outputs future sequences of the tourism demand variable.
  \item We introduce a multi-head attention CNN model (MHAC) that receives multivariable time-series inputs, extracts temporal characteristics, and interprets correlations between variables.
  \item We utilize the proposed framework to predict the demand for foreign inbound tourists in South Korea. We explore various external factors related to South Korea's tourism demand and reflect them in the forecasting model.
\end{itemize}

\section{Previous Work}

In early studies of tourism demand forecasting, research is mainly focused on predictions in specific regions and specific tourism industry sectors \cite{george2011persistence, geurts1975comparing, pattie1996using, garcia1997note, chan1993forecasting, dharmaratne1995forecasting}. Most of these studies use regression models, mainly used for time-series prediction, or classical machine learning techniques such as Support Vector Machine (SVM) \cite{hearst1998support} when designing predictive models. For instance, Chen \textit{et al.} used the support vector regression model to predict the number of foreigners entering China between 1985 and 2001 \cite{chen2007support}. Assaf \textit{et al.} devised the Detrended ARIMA model to predict the number of tourists entering Australia in both the short and long term \cite{george2011persistence}. Akın \textit{et al.} designed a prediction model based on the Seasonal ARIMA model to predict the monthly foreign inbound in Turkey \cite{akin2015novel}. 
% 관광 수요 예측과 관련한 초기 연구 예시, 주로 특정한 지역 및 국가의 관광업 또는 관광객 수요를 예측하는 연구가 주를 이룬다. %

Existing studies of tourism demand forecasting mainly focus on the discovering data from specific tourism industries of each region or country rather than designing a specific forecasting model. The regression models used as prediction models in previous studies show good performance in predicting tourism demand in particular regions. However, there are some limitations to these existing studies. First, utilized regression models such as ARMA, ARIMA, and classical machine learning models such as SVM commonly show poor forecasting performance with non-stationary time-series data. In addition, these studies constructed a prediction framework using only a single variable such as the past entrants data. There is a limitation in that various external factors affecting the tourism industry are not considered.
% 초기의 연구들은 대부분 회귀 모형이나 고전적인 기계학습 기법들을 사용했다. 이 연구들은 다른 상관성 있는 변수를 고려하지 않고 단일 변수만 고려한다는 한계점과, 회귀 모형과 SVM과 같은 기계학습 기법들이 가지는 한계점(비정상성을 띄는 데이터에서는 예측 정확도가 낮다)이 존재한다. %

Beyond the study of designing a prediction framework using single time-series data, several recent studies have proposed a multivariate prediction model using deep learning. The deep learning techniques of the recurrent neural network series such as LSTM \cite{hochreiter1997long} have received attention as a tourism demand prediction model. Zhang \textit{et al.} use historical hotel guest data, and Baidu index data to predict the trend of hotel guests in Hunan, China and designs an LSTM-based predictive model that can reflect these multiple variables \cite{zhang2019forecasting}. Kulshrestha \textit{et al.} design a prediction framework based on the Bidirectional LSTM model to predict monthly Macau visitors \cite{kulshrestha2020bayesian}. In a recent study, an LSTM model with an attention module \cite{vaswani2017attention} added is devised to consider the correlation between various input variables that influence tourism demand \cite{law2019tourism}.
%  %

\begin{figure*}[t]
\begin{center}
\includegraphics[width=1\linewidth]{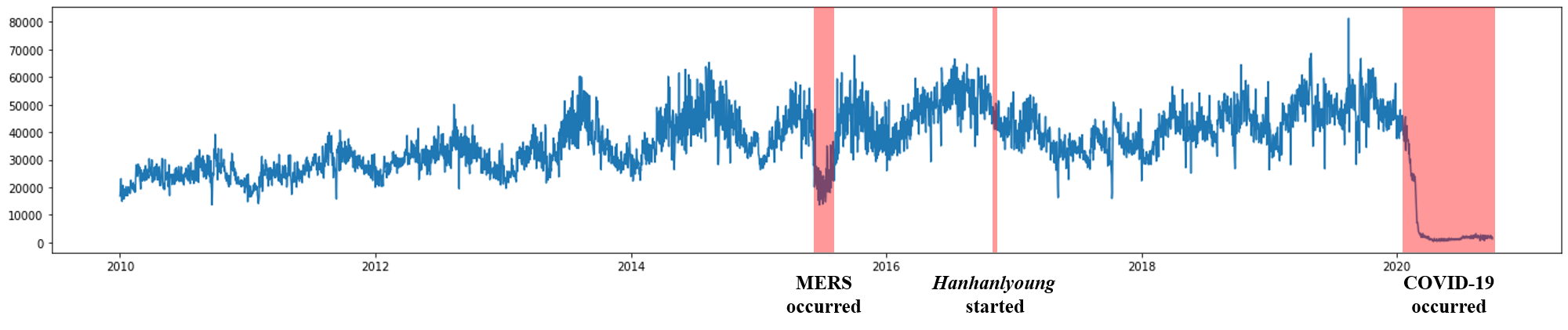}
\end{center}
   \caption{Daily data graph of foreign entrants in South Korea.}
\label{fig:fig2}
\end{figure*}

As such, recent studies attempt to analyze various time-series patterns by designing more advanced recursive models. They consider not only historical data of the variable to be predicted, but also various data related to the target variable (e.g., Google Trends, Baidu Index, etc.). However, frameworks of these studies have a structure to separately predict by constructing a recursive model for each variable. 
Since this structure cannot interpret correlations between variables, the recurrent-based models are restricted in precisely forecasting tourism demand that is affected by multiple factors. To handle this problem, a model having a structure other than the recursive model is required.
%  %

\section{Data} \label{Data}
Before explaining the forecasting framework, we introduce data to be used in forecasting South Korean tourism demand. Novel variables that haven’t been considered in the field of Korean tourism management are introduced.

\subsection{Problem Description}\label{PD}
We forecast concrete tourism demand trends for a certain period for a more practical tourism demand forecasting. The proposed prediction framework is a time-series prediction model that receives multivariate time-series data as input and outputs a sequence of a single interest variable. A multivariate time-series input \(\mathbf{X}_t\) for prediction framework is below:

\[\mathbf{X}_t = \{\mathbf{x}_{t-m+1}, \mathbf{x}_{t-m+2}, \mathbf{x}_{t-m+3}, \cdots, \mathbf{x}_t\}\]

\begin{figure}[t]
\begin{center}
\includegraphics[width=1\linewidth]{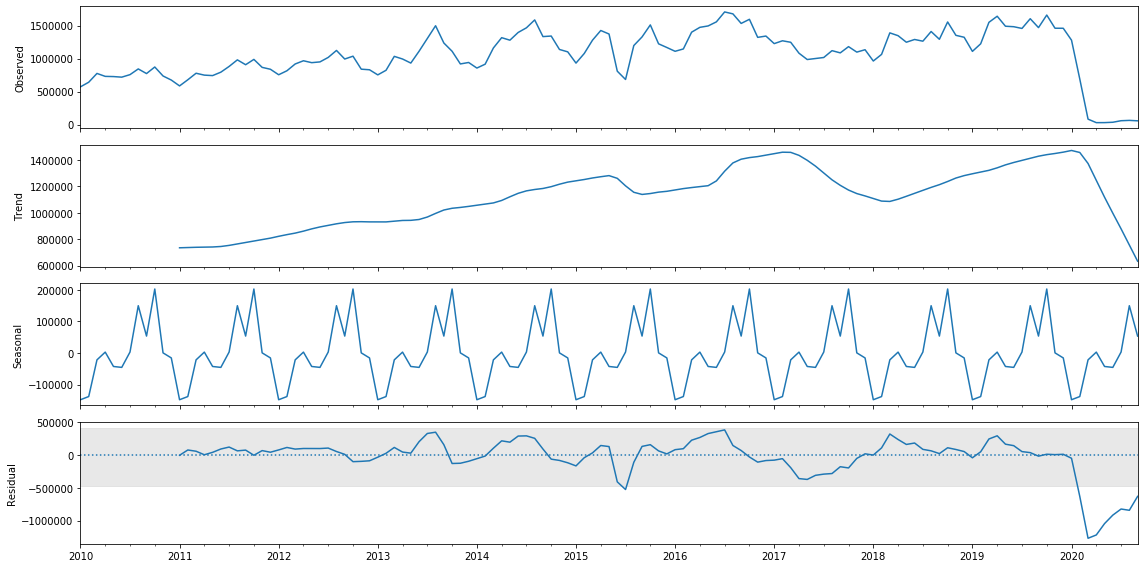}
\end{center}
   \caption{Results of time-series decomposition of processed foreign entrants data. The original data graph, trend graph, seasonality graph, and residual graph are shown from the top.}
\label{fig:fig3}
\end{figure}

\noindent where \(\mathbf{x}_i \in \mathbb{R}^{n}\) indicates \(n\) multivariate inputs at time point \(i\). Note that \(t\) refers to a specific time point, and \(m\) is the input window size. The multivariate input data having \(n\) variables which includes past foreign entrant data which is target variable, and \(n-1\) external variables related thereto. The prediction result \(\hat{\mathbf{Y}_t}\) is as follows.

\[\hat{\mathbf{Y}}_t = \{\hat{y}_{t+1},\hat{y}_{t+2}, \cdots, \hat{y}_{t+k} \}\]

\noindent Here, \(\hat{y}_i\) is the single target variable at time point \(i\), which indicates the predicted number of foreigners entering South Korea. \(k\) refers to the output window size. A ground-truth value is \(\mathbf{Y}_t\), which is shaped the same as \(\hat{\mathbf{Y}}_t\). In summary, \(\mathbf{X}_t\) is a \(n \times m\) shaped multivariate time-series input matrix, and \(\hat{\mathbf{Y}}_t\) is a \(1 \times k\) shaped univariate time-series output vector.

\subsection{Foreign Entrants Data}\label{IT}

This paper uses the foreign entrants data from January 1, 2010, to September 30, 2020, provided by the Korea Tourism Organization. The provided visitor data is aggregated in all provinces in South Korea. This data shows the daily number of foreigners entering South Korea from 21 countries, including China, Japan, and the United States. We define this foreign entrant variable as the main variable, which is a variable to use for both input and prediction. Figure \ref{fig:fig2} shows the overview of given foreign entrant data.

We would like to analyze the characteristics of this foreign entrant data and find related external factors. The time-series decomposition is performed with data on foreign entrants summed up the number of entrants before and after 15 days of a certain day (i.e., 30-day moving sum). The results of time-series decomposition of preprocessed foreign entrants data are shown in Figure \ref{fig:fig3}. 

\begin{table}[t]
\centering
\caption{A description of the input variables used in the prediction framework.}
\begin{tabular}[t]{lcccc}
\hline
Kinds&Factor&Name&Data Type\\
\hline
External&Politics&\textit{Hanhanlyeong}&Dummy Variable\\
&Disease&Pandemic&Dummy Variable\\
&Seasonality&Season&Dummy Variable\\
&Attraction&Google Trend&Numeric\\
\hline
Main&Entrant&Foreign Entrant Data&Numeric\\
\hline
\end{tabular}
\label{table:tb1}
\end{table}

The overall trend of entrants increased until 2017. As of the first half of 2017, the trend declined and then increased again. It is for the decrease in Chinese tourists, which account for the largest proportion of tourists visiting South Korea. The reason for the decline in Chinese tourists is the Chinese restrict policy against Korean culture, which is related to deteriorated Sino-South Korea Relations. Note that the number of entrants dramatically plummeted since February 2020 when the travel restrictions due to COVID-19 were taken. Like this case, the number of tourists sharply declines during the MERS epidemic (June 2015 to August 2015), which was a short-period epidemic in South Korea. The seasonality graph shows that a larger number of entrants usually comes in fall rather than in spring. The residual graph shows that the residual is very large, unlike other conventional time-series data. In conclusion of data analysis, the overall data of foreign entrants have irregular, inconsistent patterns and is influenced by certain external factors.

\subsection{Extrinsic Variables}\label{CV}

Extrinsic factors correlated to tourism should be considered as well as main variables for precise prediction. Rather than using external variables such as economy metrics or the oil price \cite{croes2005econometric, smeral2009impact}, recent studies utilize external variables directly related to forecasting tourism demand (e.g., Google trends \cite{dergiades2018google, onder2016forecasting}, seasonal components \cite{liu2018big}, climate data \cite{li2018relative}, etc.). In our method, external factors influencing South Korea's tourism demand are used as input variables.

We select external factors based on the results of the data analysis briefly mentioned above. We consider \textit{Politics}, \textit{Diseases}, \textit{Seasons}, and \textit{Attraction} as representative external variables that directly affect tourism in South Korea. A summary of the input variables considered is presented in Table \ref{table:tb1}. A description of each variable is explained below.

\noindent \textbf{Politics Variable} According to the provided foreign entrants data, Chinese tourists visit South Korea the most (About 35\%). Therefore, the demand for tourism in South Korea is greatly affected by diplomatic relations between South Korea and China. For instance, the number of foreign arrivals changes significantly because of the deterioration of Sino-South Korea relations since 2017, as mentioned above. We use the variable "\textit{Hanhanlyeong,}" a sanctions policy against Korean culture in China, as an external factor representing the diplomatic and political situation in South Korea \cite{meesak_2017}. This restriction policy has been in effect since March 2017. The \textit{Hanhanlyeong} variable is a dummy variable that consists of 0 and 1. The time-series value of this variable is set to 1 from March 1, 2017, to September 30, 2020 (Dataset Endpoint), when these sanctions policy against South Korea is in effect, and 0 for other periods.

\noindent \textbf{Diseases Variable} During the epidemic period, travel and exchanges are restricted, so the number of people entering or leaving South Korea drops sharply. From June 2015, the period when the epidemic MERS was prevalent in South Korea, to August of the same year, and from February 2020, the period when the COVID-19 pandemic worldwide, the number of foreign entrants has sharply decreased. Based on these observations, we create dummy variables that reflect the duration of the epidemic outbreak. The variable is set to 1 during the MERS outbreak period (June 1, 2015, to August 31, 2015) and the COVID-19 epidemic (February 1, 2020, to September 30, 2020), and the variable is set to 0 for other periods.

\begin{figure*}
\begin{center}
\includegraphics[width=0.75\linewidth]{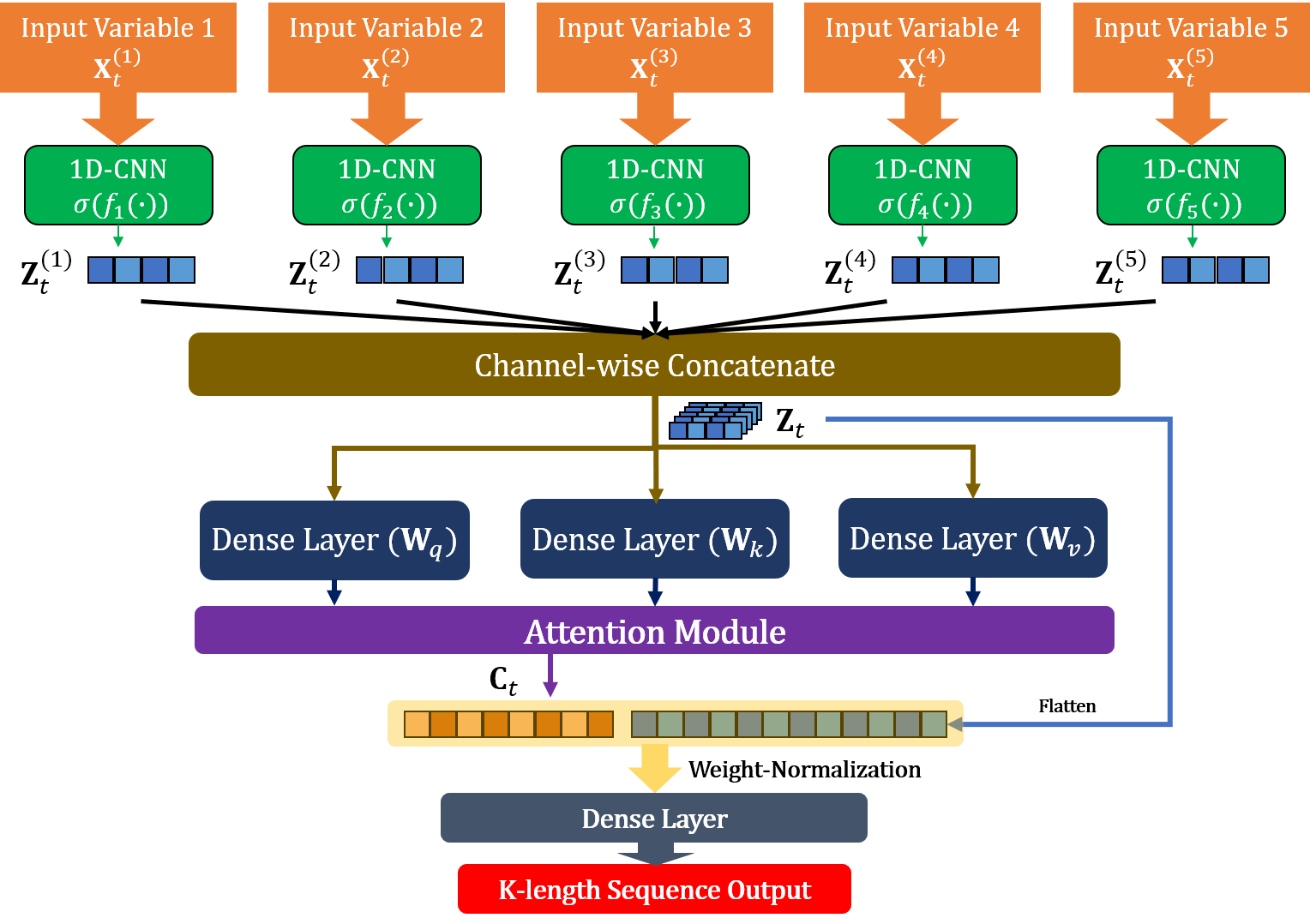}
\end{center}
   \caption{A structure of the proposed MHAC model. The time-series features of each input variable are extracted through an independent CNN layer. Then it is emphasized which part of individual features is important through the Attention Module. Finally, prediction results are derived through the weight normalization added fully-connected layer with the vector extracted through the attention module and the features extracted through the CNN layer. Note that the number notation next to the \textit{Input Variable} is the enumerated variable number.}
\label{fig:fig5}
\end{figure*}

\noindent \textbf{Seasonal Variable} As the results of the time-series decomposition in Figure \ref{fig:fig3}, entrant data has seasonality. Thus, we use seasonal dummy variables to utilize the seasonality of the data in the predictive model. The seasonal variable is a dummy variable time-series with a total of 4 channels reflecting spring, summer, autumn, and winter. The variable is set to 1 for each season and 0 for other periods.

\noindent \textbf{Attraction Variable} In several existing studies, Google Trends and Baidu Index related to each region are used in the tourism demand forecasting framework \cite{bangwayo2015can, choi2012predicting, onder2016forecasting}. It can be interpreted that searching for a search word related to tourism on a portal site indicates the degree of tourism interest in the region or country. We also consider Google Trends, which shows the trend of search volume for keywords that are highly related to Korean tourism. We choose '\textit{Seoul Hotel}', '\textit{Korea Tour}', '\textit{Incheon Airport}', and '\textit{Myeongdong}' as keywords related to Korean tourism, which represent accommodation, tourism, aviation, and attractions each \cite{park}. The Google Trends represents the relative trend of the search volume for a specific search word over a certain period of time as a real value between 0 and 100. Our work uses Google Trends data from January 1, 2017, to September 30, 2020, which is the entire period of foreign entrant data.

\section{Forecasting Model}

We would like to design a novel neural network model that is suitable for forecasting tourism demand. In this paper, a multi-head attention model with convolutional neural network layer (MHAC) is proposed. The overall structure of the predictive model MHAC is shown in Figure \ref{fig:fig5}.

\subsection{Multi-Head Structure} We introduce a multi-head neural network structure to extract features variable-wise. As described in Section \ref{Data}, the input variables have different data structures (numeric type, dummy type), and the temporal features of each variable are very diverse. Putting whole variables into a shared single neural network is unsuitable for handling multiple variables with different time-series characteristics. The proposed multi-head structure extracts the temporal features of individual variables with a parallel neural network layer. This structure has the advantage of extracting features for each variable by independently tuning the hyper-parameters for each head layer. Since there are a total of 5 types of variables, we design a forecasting model with a 5-multi-head structure.

\subsection{1-Dimensional Temporal CNN}

Several studies \cite{oord2016wavenet, cho2018divide, zhao2017convolutional, bai2018empirical} have proposed CNN-based models to process sequential data of signal processing, time-series classification, speech recognition, etc. These previous studies demonstrate a one-dimensional CNN advantage in extracting temporal features from data with irregular and diverse patterns, such as the provided foreign entrant data. We add a multi-head temporal 1D-CNN layer to interpret the pattern of the time-series data variate-wise. For the \(t\)-th input sequence \(\mathbf{X}_t^{(i)}\) of single variable \(i\), the latent feature \(\mathbf{Z}_t^{(i)}\) is extracted through the following function Equation \eqref{latent}.

\begin{equation}
\mathbf{Z}_t^{(i)} = \sigma(f_i(\mathbf{X}_t^{(i)})), \qquad i \in [1, n] \label{latent}
\end{equation}

\noindent Here, \(f_i(\cdot)\) indicates the \(i\)th head of the temporal-CNN layer, and \(\sigma(\cdot)\) refers to ReLU and MaxPool1D layer. The detailed hyper-parameters of the multi-head CNN layer are described in Section \ref{HPS}.

\subsection{Attention Module}

We add the attention module to the prediction model to reflect the correlation between each extrinsic factor and the entrant data. The attention module receives Query, Key, and Value and outputs context vector. To give attention to the extracted features \(\mathbf{Z}_t\), Query, Key, and Value are calculated through the following Equation \eqref{qkv}.

\begin{equation}
\mathbf{Q}_t = \mathbf{W}_{q}\mathbf{Z}_t, \quad \mathbf{K}_t = \mathbf{W}_{k}\mathbf{Z}_t, \quad \mathbf{V}_t = \mathbf{W}_v\mathbf{Z}_t \label{qkv}
\end{equation}

\noindent Note that \(\mathbf{W}\) indicates weight matrix for Equation \eqref{qkv}, and \(\mathbf{Q}_t\), \(\mathbf{K}_t\), \(\mathbf{V}_t\) are the query, key, and value from latent feature \(\mathbf{Z}_t\), respectively.

Then, the attention \(Score\) is obtained from Query and Key by Equation \eqref{score}.

\begin{equation}
Score(\mathbf{Q}_t, \mathbf{K}_t) = tanh(\mathbf{W}\acute{}_{q}\mathbf{Q}_t^\mathbf{T} + \mathbf{W}\acute{}_{k}\mathbf{K}_t^\mathbf{T} + \mathbf{b}) \label{score}
\end{equation}

\noindent where \(\mathbf{W}\acute{}\) refers to weight matrix for Equation \eqref{score} and \textbf{b} is bias.

Finally, Equation \eqref{context} shows how to derive the final Context vector \(\mathbf{C}_t\) through obtained \(Score(\mathbf{Q}_t, \mathbf{K}_t)\).

\begin{equation}
\mathbf{C}_t = Flatten(Softmax(Score(\mathbf{Q}_t, \mathbf{K}_t))\mathbf{V}_t^\mathbf{T}) \label{context}
\end{equation}

Context vector \(\mathbf{C}_t\) obtained through the attention layer is concatenated with the latent feature \(\mathbf{Z}_t\) and enters the input of the last fully-connected layer.

\begin{figure}[t]
\begin{center}
\includegraphics[width=0.85\linewidth]{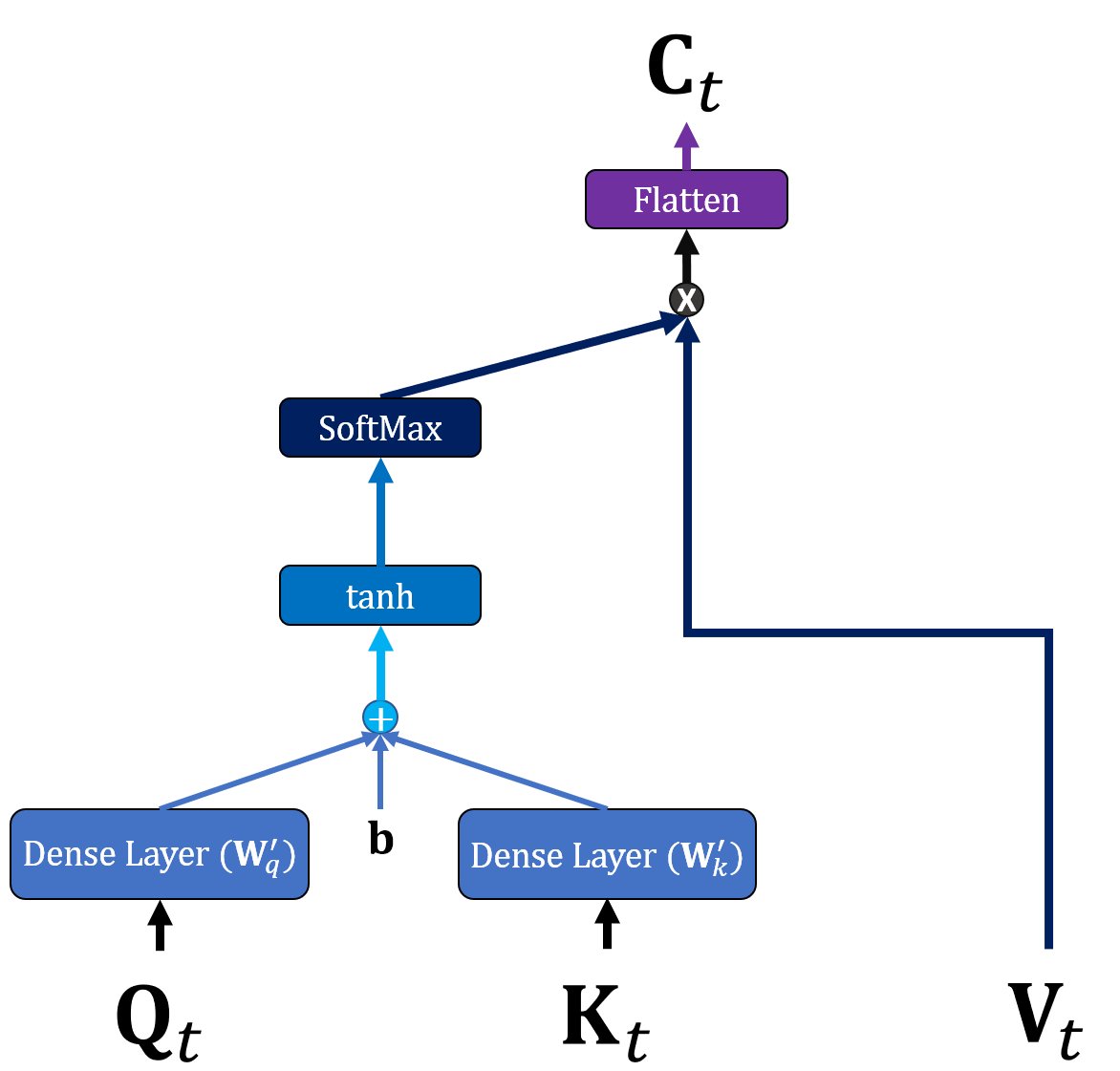}
\end{center}
   \caption{Structure of the attention module.}
\label{fig:fig6}
\end{figure}

\subsection{Weight Normalization} 

The proposed forecasting framework outputs the prediction sequence. According to Gehring \textit{et al.}\cite{gehring2017convolutional}, their experiments demonstrate that the weight normalization method achieves better performance than the conventional batch normalization method in the sequence-output framework with the CNN structure. Based on this idea, we add a weight normalization layer in front of the fully-connected layer in the forecasting model. The normalized weight vector of the fully-connected layer \(w\) is as shown in Equation \eqref{wn}.

\begin{equation}
w = \frac{g}{||v||}v \label{wn}
\end{equation}

\noindent where \(g\) is a scalar parameter, \(v\) indicates a \(k\)-dimensional vector, and \(||v||\) is the Euclidean norm of \(v\) \cite{salimans2016weight}. By adding a weight normalization layer, the proposed model can become more robust to the values of learning hyper-parameters such as learning rate. Also, weight normalization reduces the likelihood of convergence with sharp minima, thereby improving generalization performance \cite{qiao2019micro}.

\section{Experiment Setting}

\begin{figure}[t]
\begin{center}
\includegraphics[width=1\linewidth]{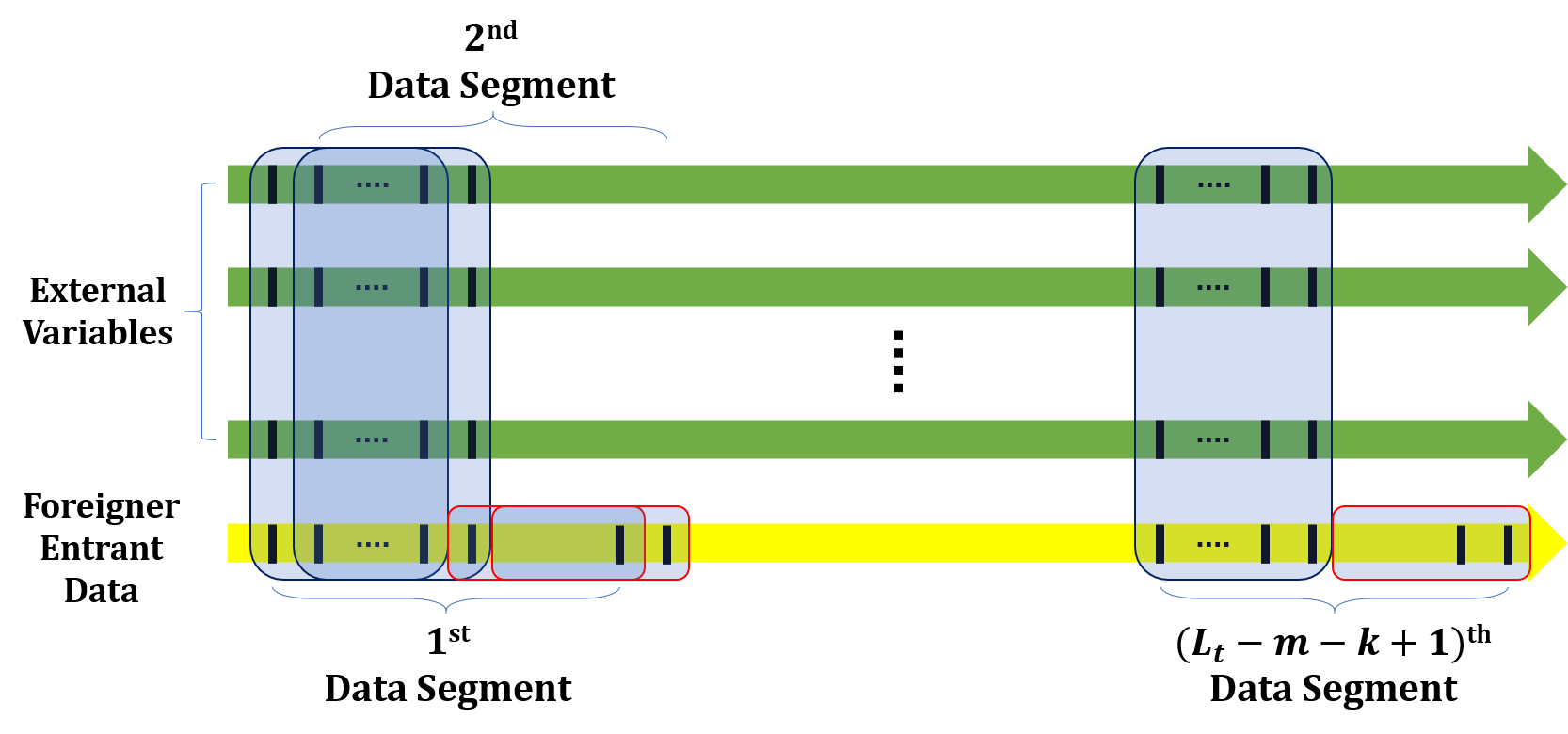}
\end{center}
   \caption{The process of creating a data segment in a sliding window manner. For a training dataset with a total length of \(L_t\), a data segment is generated by sliding the window by 1 time unit. The blue window is an input window with a length of \(m\), and the red window is a ground-truth window with a length of \(k\).}
\label{fig:fig8}
\end{figure}

\subsection{Input Data Description} 
In our experiments, the daily foreign entrant data and the daily data of the extrinsic factors are used, which are mentioned in Section III. More specifically, we use five input variables: foreign entrant data, politics (Hanhanlyeong) dummy variable, disease dummy variable, seasonal dummy variable, and attraction variable (Google trend data of keyword ’Seoul hotel’) respectively. Daily data has 3926 days from January 1, 2010, to September 30, 2020. During this period, we set data up to December 31, 2018, as training data, and data from January 1, 2019, as test data.

\subsection{Preprocessing}
As explained in Section \ref{PD}, our framework predicts the trend of entrants within the future \(k\) days through \(m\) data of \(n\) variables in the past. We set \(k = 30\), \(m = 30\), and \(n = 5\). A window of size \(m+k\) is created and data segments are generated by pushing 1 time unit for the entire dataset period using a sliding window method.  Therefore, \(L_t-m-k+1\) training data segments are generated for the total length \(L_t\) of training dataset. The first \(m\) data are input data, and the last \(k\) data are ground-truth. In the same way, test data segments are also created from the test dataset. Figure \ref{fig:fig8} shows the process of creating data segments using the sliding window method.

\subsection{Data Augmentation}

The total length of the provided daily foreign entrant data is about two years and nine months. time-series data of this length is not sufficient to train the model. Moreover, it is difficult to train with provided data since the pattern itself is very diverse compared to the data length. So, we augment the time-series data for more stable training results and higher prediction accuracy.

\begin{figure}[t]
\begin{center}
\includegraphics[width=0.8\linewidth]{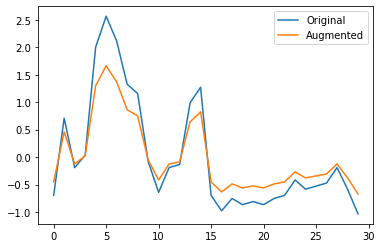}
\end{center}
   \caption{Data Augmentation Toy Example. The horizontal axis represents the time of the data segment, and the vertical axis represents the corresponding variable values. Individual variables in a data segment are augmented like this example.}
\label{fig:fig7}
\end{figure}

\begin{figure*}
\begin{center}
\includegraphics[width=0.8\linewidth]{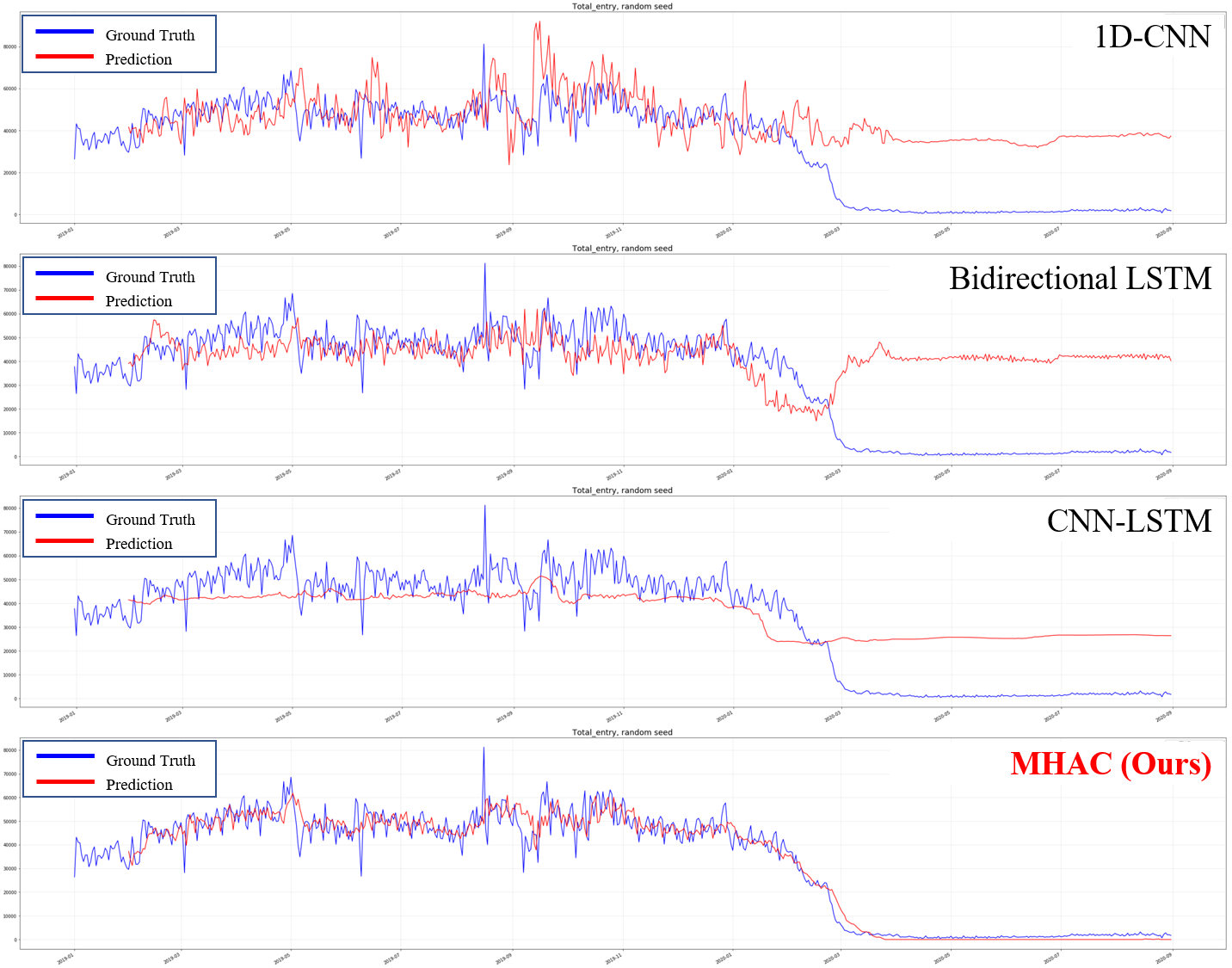}
\end{center}
   \caption{Prediction result graphs of each prediction model. From the top, 1D-CNN, Bidirectional LSTM, CNN-LSTM, and MHAC.}
\label{fig:fig9}
\end{figure*}

Unlike other augmentation methods such as cropping or rotation, time-series data is mainly augmented by using specific methods like window warping \cite{cui2016multi}, flipping, Fourier transform \cite{steven2018feature}, and down-sampling. Among them, a simple method of adding Gaussian noise is often used \cite{wen2020time}. This data augmentation technique improves performance in time-series prediction models such as DeepAR \cite{salinas2020deepar}. 

In this paper, we apply a new technique to augment time-series data to fit the proposed MHAC model. The training data is augmented by generating data obtained by multiplying statistical noise to the existing training data. The method of augmenting training data is as follows:

\begin{enumerate}
  \item Prepare variance vector \(\mathbf{V}\) for \(n\) input variables like below.
  \[\mathbf{V} = \{v_1, v_2, \cdots, v_n\}\]
  where \(v_i\) is the variance of the variable \(i\). Note that \(i\) is the numbered index of the enumerated \(n\) input variables.
  \item Prepare 1 input data matrix \(\mathbf{X}_t\) and corresponding ground-truth value \(\mathbf{Y}_t\) at time point \(t\) mentioned in Section \ref{PD}.
  \item Generate the noise \(\mathbf{\epsilon}^t_i\) for that follows the following log-normal distribution.
  \[\epsilon^t_i \sim Logn(0, 0.2 v_i)\]
  \item Create the following error matrix \(\mathbf{E}_t\).
  \[\begin{pmatrix}
  \epsilon^t_1 & 0 & \dots & 0\\
  0 & \epsilon^t_2 & \dots & 0\\
  \vdots & \vdots & \ddots & \vdots\\
  0 & 0 & \dots & \epsilon^t_n \\
  \end{pmatrix}\]
  \item Create the augmented data \(\mathbf{X}'_t\) and the corresponding ground truth \(\mathbf{Y}'_t\) through the following equations.
  \[\mathbf{X}'_t = \mathbf{E}_t\mathbf{X}_t\]
  \[\mathbf{Y}'_t = \mathbf{E}_t\mathbf{Y}_t\]
  \item Repeat for every data segment.
\end{enumerate}

Training data is augmented based on the aforementioned method. An example of augmented data is shown in Figure \ref{fig:fig7}. In this paper, the training data is augmented by a total of 9 times the training data. The performance of the model is verified only with test data that has not participated in data augmentation.

\subsection{Hyper-parameter Setting} \label{HPS}
The learning rate is 0.001, and the optimizer is Adam \cite{kingma2014adam}. The total training epoch is 50, and 20\% of the training data is used as verification data.

The number of output channels of the CNN Layer of each input variable is samely set to 4, and stride is samely set to 1. The sizes of the 1D kernel in the CNN Layers are 5, 3, 3, 3, and 5, referred to as the \textit{Foreign entrant}, \textit{Politics}, \textit{Disease}, \textit{Season}, and \textit{Attraction} variable, respectively. In CNN layers, a causal convolution is performed in consideration of the temporal features of the input variables. The same-padding is adjusted so that the length of the latent features does not change from the input shape (i.e., \(\mathbf{Z}_t^{(i)} \in \mathbb{R}^{m \times 4}\), where 4 indicates the number of output channels of the CNN Layer). The activation function of CNN Layer is ReLU. Lastly, the final fully connected layer has a 25\% dropout rate.

The batch size is 4, smaller than that of other conventional deep learning experiments because the training procedure is unstable with larger batch size. In addition, the loss function is set to Mean Squared Error (MSE).

\section{Experiment Results}
The final output result is sequential data with \(k\) lengths. Since the generated data segment was created in the form of a 1 time unit sliding window, \(k\) output results are overlapped for a single time point. We firstly use methods evaluated in the existing tourism demand forecasting study to see the concrete forecast results. So we only plot the resulting graph for a single time point. In the resulting graph to be described later, the prediction time point is after one day.

Since our forecasting framework outputs sequential forecast results, a more detailed evaluation is needed in addition to graphing the forecast results for a single time period. Therefore, we evaluate the performance of the prediction model using forecasting performance metrics \cite{shih2019temporal}. We use Mean Absolute Percentage Error (MAPE), Root Mean Squared Error (RMSE), and empirical correlation coefficient (CORR) as evaluation indicators.

Lastly, we would like to observe the reliable results of the experiments. Therefore, we conducted the same experiment 5 times and averaged the results.

\begin{table}[t]
\centering
\caption{Performance evaluation results compared to other time-series prediction models.}
\begin{tabular}[t]{c|c|c|c}
\hline
Model&MAPE&RMSE&CORR\\
\hline
1D-CNN&93.2\%&67972.5&0.3915\\

Bi-Directional LSTM&54.75\%&56871.1&0.4897\\

CNN-LSTM&102.4\%&78114.9&0.3078\\

\textbf{MHAC(Ours)}&\textbf{25.7\%}&\textbf{32195.1}&\textbf{0.7359}\\

\end{tabular}
\label{table:tb2}
\end{table}

\subsection{Comparison With Other Prediction Models}
We conduct prediction experiments with the proposed model and other deep learning models. Bidirectional LSTM \cite{ma2018parallel}, CNN-LSTM \cite{livieris2020cnn}, and 1D-CNN \cite{barra2020deep} are selected as comparison models. All of the presented comparison models are deep learning models that are frequently used in the time-series prediction field recently. Each prediction model has the same input size and output size. The prediction performance results for each model are presented in Figure \ref{fig:fig9} and Table \ref{table:tb2}.

The predictive evaluation indicators for the entire test period are presented in Table \ref{table:tb2}, which shows that the MHAC model has better predictive performance than the other deep learning models. In particular, our prediction model is superior to other prediction models in the CORR result. The MHAC model infers the overall trend better than the comparison models.

Figure \ref{fig:fig9} shows each prediction model's foreign entrants prediction results from January 1, 2019, to September 30, 2020, the test period. The blue line is the actual data, and the red line is the predicted value. According to the graphs of each prediction model, all models are somewhat inaccurate in predicting very detailed patterns—however, the proposed model, MHAC, is superior to other models in following complex patterns and trends. In particular, from February 1, 2020, to September 30, 2020, during the COVID-19 outbreak, it is observed that our predictive model infers the number of foreign entrants during this period better than other comparison models.

\begin{table}[t]
\centering
\caption{Experiment results on removing a certain external factor.}
\begin{tabular}[t]{c|c|c|c}
\hline
Variable&MAPE&RMSE&CORR\\
\hline
\textbf{With 5 variables}&\textbf{25.7}\%&\textbf{32195.1}&\textbf{0.7359}\\

w/o \textit{Politics}&36.5\%&41192.3&0.6105\\

w/o \textit{Disease}&86.7\%&60878.4&0.4017\\

w/o \textit{Season}&40.3\%&42177.9&0.5275\\

w/o \textit{Attraction}&91.9\%&65515.9&0.3990\\

\end{tabular}
\label{table:tb5}
\end{table}

\begin{table}[t]
\centering
\caption{Performance results according to the number of augmented data.}
\begin{tabular}[t]{c|c|c|c}
\hline
Augmentation&MAPE&RMSE&CORR\\
\hline
13-time Augmented&25.9\%&33585.1&\textbf{0.7360}\\

\textbf{9-time Augmented}&\textbf{25.7}\%&\textbf{32195.1}&0.7359\\

5-time Augmented&29.8\%&35988.9&0.7006\\

1-time Augmented&49.7\%&45971.7&0.6114\\

w/o Augmentation&79.9\%&60913.2&0.4389\\

\end{tabular}
\label{table:tb6}
\end{table}

\begin{table}[t]
\centering
\caption{Experiment results of tuning batch size.}
\begin{tabular}[t]{c|c|c|c}
\hline
Batch size&MAPE&RMSE&CORR\\
\hline
1&31.8\%&37005.8&0.6979\\

2&26.0\%&\textbf{32071.9}&0.7298\\

\textbf{4}&\textbf{25.7}\%&32195.1&\textbf{0.7359}\\

8&30.1\%&39267.1&0.7113\\

16&35.8\%&41936.5&0.6618\\

32&46.6\%&45991.0&0.6249\\

64&52.7\%&52107.9&0.5908\\

\end{tabular}
\label{table:tb7}
\end{table}

\begin{table}[t]
\centering
\caption{Ablation experiment results of MHAC Model.}
\begin{tabular}[t]{c|c|c|c}
\hline
Model&MAPE&RMSE&CORR\\
\hline
\textbf{Original MHAC}&\textbf{25.7}\%&\textbf{32195.1}&\textbf{0.7359}\\

w/o WN Layer&81.5\%&79681.1&0.4892\\

w/o Attention Module&51.8\%&43690.7&0.5263\\

Single CNN Structure&60.6\%&59188.6&0.4720\\

\end{tabular}
\label{table:tb4}
\end{table}

\subsection{In-Depth Experiment}
We design in-depth experiments on the proposed methods. We conduct experiments on the effect of the external factors used, the effect on data augmentation, and the batch size.

\noindent \textbf{Effect of the External Factors} We perform an experiment to compare the prediction results for whether or not external factors are used. Each experiment is performed in which a single external factor is removed. Fine-tuning is performed separately by setting the head of the MHAC model to 4. The prediction experiment results are presented in Table \ref{table:tb5}.

We observe that the prediction results when each variable is subtracted are worse than the original one. In particular, it appears that the performances are much worse when \textit{Disease} or \textit{Attraction} variable is removed. We show that our variables have some influence on the predictive performance. Also, we see that \textit{Disease} variable and \textit{Attraction} variable have a great correlation in predicting the number of foreign entrants in South Korea.

\noindent \textbf{Effect of the Data Augmentation} We design an experiment on the effect of our proposed time series data augmentation. Prediction experiments are conducted according to the number of augmented data. Experiments with 13-time augmented, 9-time augmented, 5-time augmented, 1-time augmented, and no augmentation are performed, respectively. Note that "9-time augmented" means that the entire data was augmented 9-times by the proposed method, which is for the original experiment. The forecasting results for each experiment are shown in Table \ref{table:tb6}.

As shown in the experimental results, we observe that the performance improves as the number of data increases. However, there is no significant difference in the performance between the 9-time and 13-time augmentation experiments. It is assumed that as the augmented data becomes too large, data redundancy occurs and the performance is not significantly improved.

\noindent \textbf{Study about Batch Size} We design experiments with different batch sizes. Experiments on the batch size are performed for 1, 2, 4, 8, 16, 32, and 64, respectively. The performance results of tuning batch size are presented in Table \ref{table:tb7}.

As we described in Section \ref{HPS}, the prediction accuracy drops sharply as the number of batches increases. On the one hand, there is no significant improvement in prediction performance when the batch size is extremely small (1, 2). We see that the performance is sensitively dependent on the batch size in our framework.

\subsection{Ablation Studies on the Structure} We additionally conduct ablation experiments to determine how much the weight normalization, attention module, and multi-head structure in the MHAC model affect the prediction results.

\noindent \textbf{Weight Normalization} We would like to analyze how much weight normalization (WN) affects the prediction performance. So an ablation experiment is performed by removing weight normalization from the fully connected layer of the proposed MHAC model. All other settings are the same. The experimental ablation results are presented in Table \ref{table:tb4}.

In the results of Table \ref{table:tb4}, the predictive model's performance is worsened when weight normalization is removed. Especially, the RMSE metrics differ greatly because weight normalization allows the forecasting model to interpret detailed patterns better, while the MHAC model without weight normalization shows a large error due to unstable learning.

\noindent \textbf{Attention Module} We compare the original MHAC model and the MHAC model with only the attention module removed. All other conditions remain the same except for the attention module between the convolutional and fully connected layers. The prediction results for the ablation experiment of the attention module are presented in Table \ref{table:tb4}.

The results when the attention module is included in the predictive model are better than when the attention module is not included. Extracting the correlation between each variable through the attention module of the features gathered through the convolutional layer helps in a good forecasting result.

\noindent \textbf{Multi-Head Structure} The experiment is conducted by replacing the CNN part of the MHAC model with the multi-head CNN structure into a single CNN layer. Input variables of the same input size are input to the model channelwisely. The attention module and the fully connected layer of the prediction model are the same. The experimental results for the multi-head structure are presented in Table \ref{table:tb4}. 

As shown in the results of Table \ref{table:tb4}, the multi-head CNN structure and the single CNN structure show very large differences in the prediction results. In addition, the prediction performance differs greatly in the CORR metric, which shows the multi-head CNN structure is more advantageous in inferring the long-term trend than the single CNN structure.

\section{Conclusion}

We propose a multi-head attention CNN model to predict the foreign entrants of South Korea with high accuracy. Not just using only past foreign entrants data, we additionally utilize various external factors related to Korean tourism as input variables in the prediction model. By conducting a comparative experiment with other deep learning prediction models, it is shown that the proposed prediction model has higher accuracy in predicting South Korea’s tourism demand.
As a future study, we will explore whether the proposed model can be used for forecasting tourism demand in other countries and for other multivariate time series forecasting fields.

\section*{Acknowledgment}
This work was partly supported by Institute of Information \& communications Technology Planning \& Evaluation (IITP) grant funded by the Korea government(MSIT) (IITP-2021-2015-0-00742, Grand Information Technology Research Center support program, 50\%) and Korea Tourism Organization (A Prior Study of Foreign Tourists Prediction Model Using Artificial Intelligence, 50\%)

\bibliographystyle{ieeetr}
\bibliography{IEEEbib}

\end{document}